\documentclass[letter,10pt,conference]{IEEEtran}
\IEEEoverridecommandlockouts
\usepackage{cite}
\usepackage{amsmath,amssymb,amsfonts}
\usepackage{graphicx}
\usepackage{textcomp}
\usepackage{xcolor}
\usepackage{tabularx}
\usepackage{color, colortbl}

\definecolor{lime}{rgb}{0.88,2,10}
\usepackage[left=0.68in, right=0.6in, top=0.6in, bottom=0.68in]{geometry}
\renewcommand{\baselinestretch}{.945}
\usepackage{subcaption}
\usepackage{wrapfig}
\usepackage[linesnumbered,ruled,vlined]{algorithm2e}
\usepackage{xpatch}
\usepackage{url}
\usepackage{multirow}
\usepackage{multicol}
\usepackage{wrapfig}

\usepackage[export]{adjustbox}
\usepackage[numbers,sort&compress]{natbib}
\usepackage{nomencl}
\usepackage{siunitx}
\usepackage{blindtext}
\usepackage[T1]{fontenc}
\usepackage[spaces,hyphens]{xurl}
\usepackage{float}

\SetAlFnt{\small}

\SetKwComment{Comment}{$\triangleright$\ }{}

\usepackage[utf8]{inputenc}
\usepackage{enumitem}

\SetKwInput{kwInit}{Initialize}

\def\BibTeX{{\rm B\kern-.05em{\sc i\kern-.025em b}\kern-.08em
    T\kern-.1667em\lower.7ex\hbox{E}\kern-.125emX}}

\usepackage[numbers,sort&compress]{natbib}
\usepackage[font={small}]{caption} 

\newcommand{\sref}[1]{Section~\ref{#1}}
\usepackage[misc]{ifsym}

\begin{document}
\title{FedHAP: Fast Federated Learning for LEO Constellations using Collaborative HAPs\vspace{-0.2cm}}

\author{Mohamed Elmahallawy and Tie Luo\textsuperscript{\Letter}\\
{Computer Science department, Missouri University of Science and Technology, USA}\\
{E-mail: \{meqxk, tluo\}}@mst.edu\vspace{-0.6cm}}

\maketitle

\begin{abstract}
Low Earth Orbit (LEO) satellite constellations have seen a surge in deployment over the past few years by virtue of their ability to provide broadband Internet access as well as to collect vast amounts of Earth observational data that can be utilized to develop AI on a global scale. As traditional machine learning (ML) approaches that train a model by downloading satellite data to a ground station (GS) are not practical, Federated Learning (FL) offers a potential solution. However, existing FL approaches cannot be readily applied because of their excessively prolonged training time caused by the challenging satellite-GS communication environment. This paper proposes FedHAP, which introduces high-altitude platforms (HAPs) as distributed parameter servers (PSs) into FL for Satcom (or more concretely LEO constellations), to achieve fast and efficient model training. FedHAP consists of three components: 1) a hierarchical communication architecture, 2) a model dissemination algorithm, and 3) a model aggregation algorithm. Our extensive simulations demonstrate that FedHAP significantly accelerates FL model convergence as compared to state-of-the-art baselines, cutting the training time from several days down to a few hours, yet achieving higher accuracy. 
\end{abstract}
\vspace{-0.2cm}
\section{Introduction}\label{Intro}

Low Earth Orbit (LEO) satellite constellations are a collection of small satellites orbiting in space between 500 and 2000 km above the Earth's surface. 
Due to the low altitude and the resulting lower cost and complexity, they have recently been increasingly deployed in space to enable large-scale data collection (e.g., of Earth observation imagery) in a wide range of applications such as urban planning, weather forecast, climate change, and disaster management \cite{perez2021airborne}. To derive valuable insights from these data, machine learning (ML) offers a powerful tool but traditional approaches are no longer practical, as they download massive satellite data to a ground station (GS) to train a central ML model. Thus, there will face serious problems of limited bandwidth, intermittent and irregular satellite-GS connectivity, and data privacy \cite{abbas2021hybrid}.

Federated learning (FL) \cite{fl2021} emerged recently and appears to be a promising solution, in which each client (satellite in our case) trains an ML model locally without uploading its data anywhere, and only transmits the trained model parameters to a parameter server (PS, which is typically a GS). The PS then aggregates the received model parameters into a global model and sends it back to all the satellites again for re-training. This procedure repeats iteratively until model convergence. 

\textbf{Challenges.} However, applying FL to satellite communication (Satcom) faces several challenges. First, the connectivity or ``visibility'' of each satellite to the GS is highly intermittent and irregular, due to the distinction between satellite trajectories and Earth rotation. In fact, the interval between two consecutive visits of a satellite to the GS can vary from a couple of hours to more than a day \cite{so2022fedspace,razmi}. This will result in a huge convergence delay, up to several days \cite{chen2022satellite}, due to the iterative nature of FL. 
The second challenge is that the wireless channels between satellites and GS are highly unpredictable and unreliable as they are constantly impacted by weather conditions (e.g., rain, fog, wind turbulence) and radio interference, which are especially notable near the Earth's surface. Third, Satcom is subject to long propagation and transmission delays, due to the long communication distance and low data rate. 

\textbf{Related work.} Recently, FL has drawn intense interest due to its promising prospects in LEO constellations \cite{chen2022satellite, razmi2022ground, so2022fedspace, razmi}. Chen et al. \cite{chen2022satellite} demonstrate some benefits when applying the original FedAvg \cite{mcmahan2017communication}, unchanged, to Satcom, as compared to traditional centralized ML. Razmi et al. \cite{razmi} attempted to reduce the long training time by proposing FedISL, which employs inter-satellite-link (ISL) to improve performance. However, in order to address the poor visibility, they assumed a medium Earth orbit (MEO) satellite orbiting above the Equator to serve as the PS, which is hardly available. In addition, the {\em Doppler effect} greatly amplified by the large speed difference between MEO and LEO satellites is also overlooked by the study. 
Another work \cite{razmi2022ground} proposes FedSat as an asynchronous FL algorithm to reduce the training delay, but it assumes an ideal setup where the GS is located at the North Pole (NP) so that each satellite will visit the GS at regular intervals periodically. So et al. \cite{so2022fedspace} developed a semi-asynchronous FL algorithm called FedSpace to balance the idleness in synchronous FL and staleness in asynchronous FL when applied to Satcom. FedSpace takes an approach similar to FedBuff \cite{nguyen2022federated}, but outperforms FedBuff by additionally designing an algorithm for scheduling the aggregation process to take satellite connectivity into account. However, FedSpace requires each satellite to upload a portion of its raw data to the GS, which contradicts the desirable FL principles on communication efficiency and data privacy.

\textbf{Contributions.} In this paper, we introduce high altitude platforms (HAPs) into FL to act as PSs to orchestrate the training process for LEO constellations. Based on that, we propose a synchronous FL framework called FedHAP to achieve both {\em low training delay} and {\em high model accuracy} simultaneously. The main idea is to explore inter-satellite and inter-HAP collaborations using a novel {\em model dissemination} algorithm and {\em partial model aggregation}, under a hierarchical communication architecture (as opposed to the star architecture in conventional FL). A HAP is a semi-stationary aerial station that floats in the stratosphere at 18-24 km above the Earth's surface \cite{xing2021high,hsieh2020uav} and is equipped with communication and computing facilities. Compared to GSs or 
MEO satellites, HAPs have the following advantages: 1) {\em lower cost}---a GS or an MEO satellite typically costs over a million dollars 
while a HAP costs only a small fraction of it \cite{hap,kurt2021vision}; 2) {\em improved visibility}---a HAP will see more satellites at a time or see each satellite more often (than a GS) thanks to its much-elevated altitude; 3) {\em better communication environment}---the space in and above the stratosphere is much clearer, stabler, and less interfered than the troposphere (right above the ground); 
4) {\em easy relocation}---HAPs can move in response to changes in the LEO constellation (e.g., additional satellites and orbits are launched or existing ones retire), thereby maintaining good visibility of satellites.

However, introducing HAPs into FL-Satcom is non-trivial and calls for new architecture and algorithm redesigns. 
Correspondingly, this paper makes the following contributions:
\begin{itemize}[leftmargin=*]
    \item We introduce HAPs in lieu of GSs to act as PSs in FL to train ML models collaboratively with satellites, in a multi-orbit LEO constellation to achieve fast convergence.
    \item We propose a FedHAP framework that consists of three components: (i) a hierarchical non-star communication architecture, (ii) a model dissemination algorithm that overcomes the challenge of sporadic satellite-HAP visits, and (iii) partial model aggregation that accelerates global model convergence.
    \item We evaluate the performance of FedHAP in a wide range of settings (IID vs. non-IID, CNN vs. MLP, GS vs. HAP, single HAP vs. multi-HAP) with multiple state-of-the-art FL-Satcom approaches. The results show that FedHAP significantly outperforms them on both convergence speed and model accuracy.
\end{itemize}

\vspace{-3mm}
\section{System Model} \label{section 1}
\begin{figure} 
     \centering
     \includegraphics[width=0.9\linewidth,height=7cm]{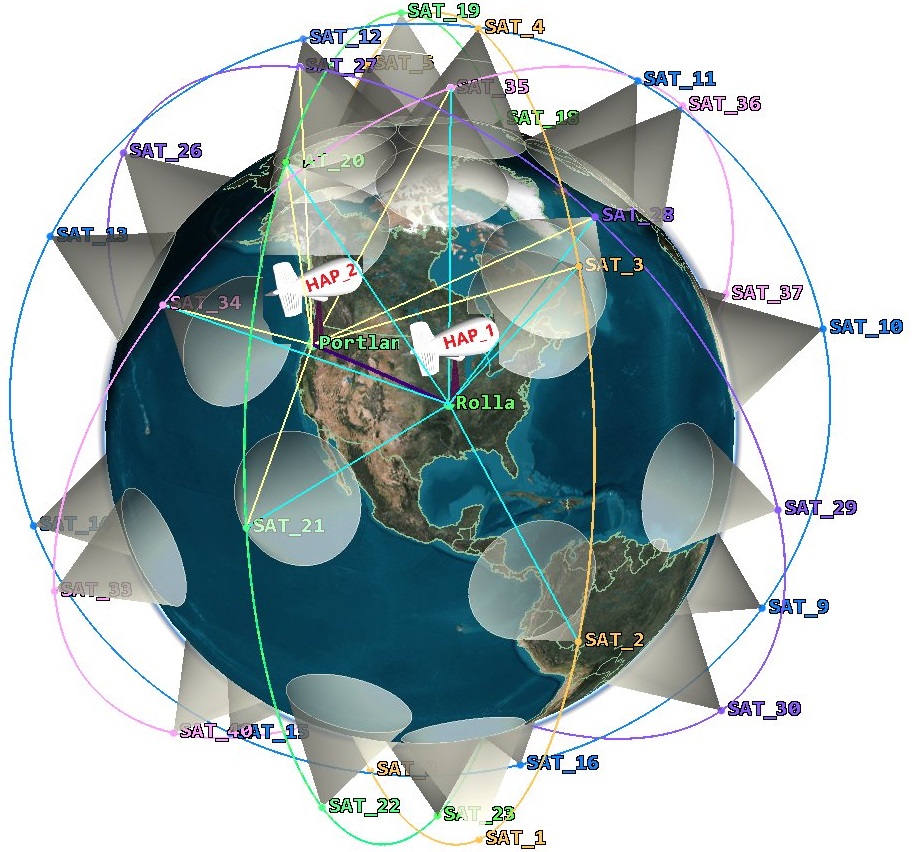}
    \caption{A {\em Walker} constellation \cite{walker1984satellite}
    consisting of $L = 5$ orbits, each carrying $K_{l}= 8$ satellites. The constellation is orchestrated by $H = 2$ HAPs. Each gray cone depicts the covered area of a satellite. }
    \label{Picture1}
    \vspace{-0.5 cm}
\end{figure}

\vspace{-0.1cm}
Consider an LEO constellation that consists of ${L}$ orbits, where each orbit $l$ carries $K_{l}$ equally-spaced satellites. 
Each satellite has a unique ID, and travels on orbit $l$ with speed $v_{l}={\frac{2\pi{(R_{E}+h_{l}})}{T_{l}}}$, where $R_{E}$ is the radius of the Earth, $h_{l}$ is the orbital height, and $T_{l}$ is the orbital period given by $T_{l}= \frac{2 \pi}{\sqrt{GM}}{(R_{E}+h_{l})^{3/2}}$, where $G$ is the gravitational constant and $M$ is the mass of the Earth.
In addition, there are $H$ HAPs where each HAP $h$ acts as a PS that communicates with a varying number of satellites from different orbits at any given time, and perform (partial or full) model aggregation for FL. An illustration is given in Fig.~\ref{Picture1}.
\vspace{-0.2cm}

\subsection{Federated learning in LEO Constellations}

For ease of description, in this section we assume a single PS as in classical FL. Later in Section \ref{sec:fedhap}, we propose a more sophisticated scheme that involves multiple HAPs acting as PSs and a new communication (non-star) architecture.

Given an LEO constellation $\mathcal K$, the overarching goal of the PS and all the satellites $k\in \mathcal K$ is to collaboratively solve the following optimization problem:
\vspace{-0.1cm}
\begin{equation} \label{eqn1}
           \displaystyle \arg \min_{w \in  \mathbb{R}^{d}} F(w)= \sum_{k\in \mathcal K }{\frac{n_{k}}{n}{F_{k}{(w)}},}
\end{equation}
where $w$ is the target ML model (i.e., its parameters), $n_{k}$ is the data size of satellite ${k}$, ${n} = \sum_{i=k}^{\mathcal K} n_{k}$ is the total size, and $F_{k}$ is the loss function that satellite $k$ aims to minimize, which is defined by 
\vspace{-0.25cm}
\begin{equation}
     F_{k}{(w)} = \frac{1}{n_{k}}\sum_{x\in D_{k}} l(w;x),
\vspace{-0.2cm}
\end{equation}
where $l(w;x)$ is the training loss over a data point \textit{x} and $D_{k}$ is satellite $k$'s local dataset whose size is $n_{k}=|D_{k}|$.

To solve the above problem, the PS first creates an ML model (e.g., a neural network) with initial parameters $w^0$ and disseminates it to all or a subset of the satellites when they (successively) come into its visible zone. Each satellite $k$ then applies a local optimization method such as mini-batch gradient descent for $I$ local epochs, to update the model as
\begin{equation}
    w_{k}^{\beta,i+1} = w_{k}^{\beta,i}- \zeta \nabla F_{k}(w_{k}^{\beta,i};X_{k}^{i}),\quad i=0,1,2,..,I-1, 
\end{equation}
where  $w_{k}^{\beta,i}$ is the local model of satellite $k$ at local iteration $i$ in a global communication round $\beta$, $\zeta$ is the learning rate, and $X_{k}^{i}\subset D_{k}$ its the $i$-th mini-batch. Once the PS receives the updated parameters from those satellites, it aggregates them as
\vspace{-0.2cm}
\begin{equation} w^{\beta+1} = \sum_{k\in \mathcal K} \frac{n_{k}}{n} w_{k}^{\beta,I}, \beta=0,1,2,..
\vspace{-0.15cm}
\end{equation}
In other words, the PS starts a new round $\beta+1$ by transmitting the updated $w$ to all the satellites again when they become visible, and the above procedure repeats until the FL model is converged (e.g., a target loss or accuracy is achieved)

There are two general FL approaches. \textbf{\em Synchronous FL}, 
such as proposed in \cite{mcmahan2017communication}, is similar to what is described above, in which the PS must wait until all the selected satellites to finish their local training and send their trained models back to the PS before moving on to the next round. \textbf{\em Asynchronous FL}, as suggested in \cite{xie2019asynchronous}, allows the PS to proceed earlier to the next round with only a (varying) subset of the satellites' training results, to avoid bottleneck or straggler satellites with limited visibility to the PS. Although this sounds appealing, it does not necessarily lead to better performance, because it requires a careful trade-off between the reduced waiting time (desired) and less training progress (undesired) in each round, as well as handling {\em stale} model parameters from {\em straggler} satellites. 

FedHAP is a synchronous FL approach and we will compare it with both synchronous and asynchronous state-of-the-art methods.

\vspace{-2mm}
\subsection{Communication Links in LEO constellations}\label{Com_link}
\vspace{-1mm}

We consider the following types of communication links: 1) ISL, 2) satellite-HAP link (SHL), 3) inter-HAP link (IHL), and 4) satellite/HAP-ground link (S/HGL). 
The communication links between satellites and the GS are radio frequency (RF) links which are full-duplex, while between satellites and HAPs are free-space optical (FSO) links which are half-duplex.

\subsubsection{RF Links}
In order to compare our proposed approach that uses HAPs as PSs to the state-of-the-art that uses GS as a PS, we used RF as a communication link thanks to their reliability for long-range communication. 
Without loss of generality, let us consider a satellite \textit{k} and a GS \textit{g}, where they will only be feasible to each other (i.e, establish a SGL between them for communicating) if the following condition is satisfied:  $\angle (r_{g}(t), (r_{k}(t) - r_{g}(t))) \leq \frac{\pi}{2}-\alpha_{min}$ where $r_{k}(t)$ and $r_{g}(t)$ are the trajectories of satellite \textit{k} and GS \textit{g}, respectively, and $\alpha_{min}$ is the minimum elevation angle. In addition, since the wireless channel in free space is AWGN (additive white Gaussian noise), the signal-to-noise ratio (SNR) between any two objects {\em a} \textit{and} \textit{b} (e.g., satellite and GS or HAP) can be written as:
\vspace{-0.15cm}
\begin{equation}
    SNR_{RF} = \frac{P_{t}G_{a}G_{b}}{K_{B} T B \mathcal{L}_{a,b}},  
\end{equation}
where $P_{t}$ is the power transmitted by the sender, and $G_{a}, G_{b}$ are the total antenna gain of the sender and receiver, respectively, $K_{B}$ is the Boltzmann constant, \textit{T} is the noise temperature at the receiver, \textit{B} is the channel bandwidth, and $\mathcal{L}_{a,b}$ is the free-space pass loss, which can be given by:
\vspace{-0.05cm}
\begin{equation}
    \mathcal{L}_{a,b} =
      \Big(\frac{4\pi \|a,b\|_{2} f}{c}\Big)^{2}, \quad\quad\quad  \text{when}\|a,b\|_{2} \leq \mathnormal{\ell_{a,b}}, 
\end{equation}
where $\|a,b\|_{2} $ is the Euclidean distance between \textit{a} and {\em b}, $f$ is the carrier frequency, and $\mathnormal{\ell_{a,b}}$ is the minimum distance between \textit{a} and {\em b} that enables them to communicate with each other (i.e, a line-of-sight link (LoS) exists between them). In other words, the visibility between the objects \textit{a} and {\em b} will be obstructed by the Earth if $\|a,b\|_{2} >\mathnormal{\ell_{a,b}}$. The overall time delay $t_{d}$ of the communication link between objects \textit{a} and {\em b} can be formulated as:
\vspace{-0.2cm}
\begin{equation}
 t_{d} = {\underbrace{\frac{z{|\mathcal D|}}{R}}_{t_{t}}}+\underbrace{\frac{\|a,b\|_{2}}{c}}_{t_{p}}+t_{a}+ t_{b}, 
    \label{eqny} 
    \vspace{-0.15cm}
\end{equation}
where $t_{t}$ \textit{and} $t_{p}$ are the transmission and propagation delay between the sender and receiver, respectively, $t_{a}$ \textit{and} $t_{b}$ are the processing delay at $a$ and $b$ when one of the objects acts as a sender, and the other as a receiver, $|\mathcal D|$ is the number of data samples, $z$ is the bits number of each sample, and $R$ is the data rate, which can be approximately calculated using the Shannon formula:
\vspace{-0.3cm}
\begin{equation} \label{eqnz}
   \quad\quad  R \approx B \log_{2} (1+SNR)
\end{equation}

\subsubsection{FSO Links}
FSO is more suitable for short-range communications due to its higher data rates and resistance to interference. HAPs operate as PSs and communicate with each other or with satellites via FSO (but we remove this difference in our simulation for a fair comparison).
Consider again two objects  $a$ and $b$ (e.g., satellites or HAPs), one as a receiver equipped with photodetectors and the other as a sender equipped with light-emitting diodes (LEDs). When an LoS optical link is established between $a$ and $b$, the channel gain can be expressed as \cite{saeed2021point}:
\begin{equation}
     {G} = \frac{\sigma+1}{2\pi (||a,b||_2)^{2}}\mathcal{A}_{0} \cos^{\sigma}({\alpha_e })T_{f} g(\theta )\cos(\theta),
\end{equation}
where $\sigma$ is the radiation coefficient (Lambertian emission), $\mathcal{A}_{0}$ is the active area of the photodetector, $\alpha_e$ is the viewing angle, $T_{f}$ is the filter transmission coefficient, $g(\theta)$ is the concentration gain, and $\theta$ is the incident angle at the receiver axis. Therefore, the SNR at the receiver can be calculated as follows:
\vspace{-0.15cm}
\begin{equation}
    \quad\quad  SNR_{FSO} = \frac{{(\rho \hspace{0.1cm}{G} P_{t}})^{2}B}{N R},
\end{equation}
where $\rho$ is the responsibility of the photodetector, and \textit{N} represents the noise variance. 
When the light is spread due to changes in the atmosphere, the receiver collects less power, resulting in poor communication. This detects as a geometrical loss at the receiver, and can be calculated as follows:
\begin{equation}
   {l}_{g} = \frac{4 \pi r^{2}}{\pi (\xi\|{a,b}\|_{2})^{2}},
\end{equation}
where $r$ is the receiver aperture radius and $\xi$ is the divergence angle. Turbulence loss is another loss where its intensity is based on the refractive index parameter $\mathcal{M}^{2}(z)$, which can be estimated using the Hufnagel-Valley (H-V) model as follows:
\vspace{-0.1cm}
\begin{equation}
\begin{aligned}
    \mathcal{M}^{2}(z) = 0.00594 \Big(\frac{\mathcal{V}}{27}\Big)^{2} \Big(10^{-5} z\Big)^{10} \exp\Big(\frac{-z}{1000}\Big) \\+2.7\times10^{-16}\exp\Big(\frac{-z}{1500}\Big)+ \mathcal{K} \exp\Big(\frac{-z}{100}\Big),
    \end{aligned}
\end{equation}
where $\mathcal{V}$ is the wind speed, $z$ is the object altitude, and $\mathcal{K} $ is constant ($1.7 \times 10^{-14}m^{-2/3}$). Once the value $\mathcal{M}^{2}(z)$ is estimated, the turbulence loss ${l}_{t}$ can be determined as follows:
\begin{equation}
    {l}_{t} =  \sqrt[2]{23.17 \Big(\frac{2 \pi f}{c}10^{9}\Big)^{7/6}\mathcal{M}^{2}(z)\big(\|{a,b}\|_{2}\big)^{11/6}}
\end{equation} 

In our simulation (Section~\ref{section 3}), the parameters for the formulas presented above will be set. 

\section{FedHAP Framework} \label{sec:fedhap}
FedHAP is a synchronous FL framework tailored to Satcom to accelerate the convergence of FL and improve its accuracy. FedHAP addresses two main challenges: (i) sporadic connectivity between satellites and PS, which causes a long convergence time in traditional FL approaches (e.g., FedAvg \cite{mcmahan2017communication}); (ii) the large number of communication rounds typically required for FL convergence. By introducing HAP as PS, the number of visible satellites at a time by a PS is increased (a GS can only ``see'' an angular range of $180^o - \alpha$, while a HAP can ``see'' even beyond $180^o$). In addition, FedHAP proposes  three new components as mentioned in \sref{Intro} that leverage inter-satellite and inter-HAP collaborations, resulting in an accelerated convergence of the global FL model. 




\subsection{FedHAP Communication Architecture}
We consider a hierarchical communication architecture consisting of two tiers. The first tier is the {\em worker tier}, which is comprised of all the satellites of the LEO constellation $\mathcal K$ that train and transmit local models using ISL or SHL. The second tier is the {\em server tier}, which is comprised of all the HAPs that aggregate and transmit global models using IHL or SHL. It is worth noting that each satellite has four antennas, two on the {\em roll axis} for intra-plane ISL communications, and two on the {\em pitch axis} for inter-plane ISL communications. As the latter is strongly affected by the {\em Doppler shift}, we only use the former (intra-plane ISL) and refer to it as ISL for short in the rest of the paper unless otherwise specified.

In the traditional FL communication architecture, i.e., star, each satellite individually communicates with the PS and, thus, the PS has to wait for each satellite to come into its visible zone, causing a significant delay even for one round of collecting model updates. Instead, we lay a {\em point-to-point (PTP) communication structure} on each orbit in the worker tier, in which only one visible satellite with a long visibility window will connect to the server tier (detailed in the next subsection).
In the server tier, the HAPs are organized in a ring architecture to facilitate communication among them (explained in the following subsection), but each HAP also communicates with a set of satellites from different orbits, forming multiple PTP connections. The result is a {\em ring of multiple PTP architectures}.

This hierarchical communication structure significantly enhances the continuity of the satellite-PS connectivity via parallel communications among rings and among multiple PTPs. 
Even when there is only a single HAP, concurrent rings can still reap substantial efficiency gains since satellites do not have to wait hours to send or receive models in their next visit to the PS. Instead, they can leverage the current or soon-to-be visible satellite to exchange models.
Additionally, all links except GS (IHL, SHL, and ISL) benefit from FSO links which have higher data rates than RF links (though we do not exploit this benefit in our experiments to ensure a fair comparison with baselines). 

\subsection{Dissemination of Local and Global Models} \label{sec:relay}

We propose a model dissemination algorithm that disseminates local and global models within and between the worker and server tiers. The main objective of this algorithm is to minimize the {\em idleness} existing in traditional synchronous FL approaches, where a PS has to idly wait for all the invisible satellites to become visible (successively) to exchange model updates. To this end, we allow each visible satellite to send both its local model and the global model (received from the server tier) to its next-hop invisible satellite which will perform partial aggregation (cf. \eqref{eq:partialagg}, explained later), until reaching the next visible satellite, thereby reducing idle waits significantly. Our dissemination algorithm consists of three phases, which are discussed below.

\subsubsection{Inter-HAP Dissemination of Global Models}\label{sec:interhap}
\begin{figure}[ht]
     \centering
     \vspace{-.25cm}
     \includegraphics[width=1.0\linewidth]{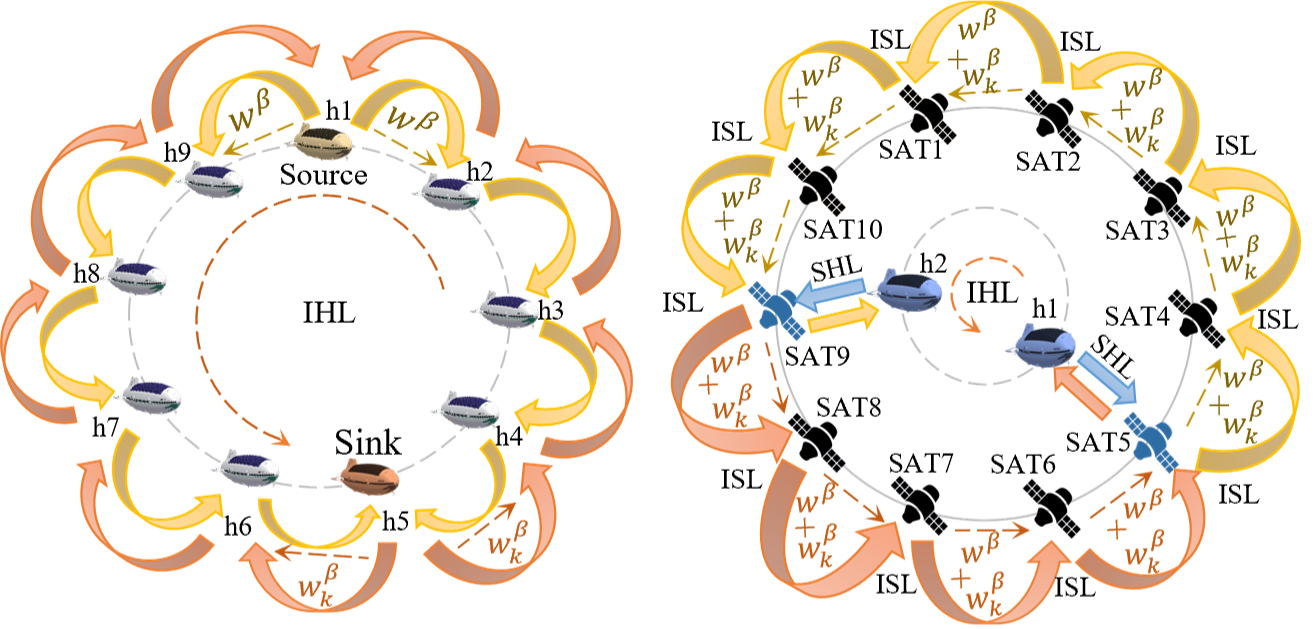}
     \begin{minipage}[t]{.5\linewidth}
       \centering
        \subcaption{\footnotesize{Inter-HAP model dissemination.}}
      \end{minipage}
      \begin{minipage}[t]{.5\linewidth}
          \centering
           \subcaption{\footnotesize {Inter-satellite model dissemination.}}
      \end{minipage}
           \vspace{-.25cm}
    \caption{Illustration of the proposed model dissemination algorithm. In (a), a source HAP ($h1$) sends a {\em global model} to a sink HAP ($h5$), indicated by yellow arrows; later on, the sink HAP will send a {\em partial-global model} to the source HAP, as indicated by orange arrows. In (b), blue satellites represent visible satellites while black ones represent invisible satellites, to HAPs $h1$ and $h2$; yellow and orange arrows represent the dissemination of models (each satellite's local model and an updated partial global model) from SAT5 and SAT9, respectively.}
    \label{Picture2}
    \vspace{-0.2cm}
\end{figure}

It is carried out at the server tier only when there are multiple HAPs. One HAP is pre-designated as the {\em source} and another (e.g., the one farthest from the source) as the {\em sink}. The source HAP generates an initial global model, $w^{0}$, and then transmits it to its adjacent HAPs via IHL. It also transmits $w^{0}$ to all of its currently visible satellites via SHL. Upon receiving $w^{0}$, each HAP sends $w^{0}$ to its next-hop neighbor on the server-ring architecture and also transmits $w^{0}$ to all of its currently visible satellites via the multi-PTP architecture, similarly to the source HAP. This continues until the sink HAP receives the model and transmits it to its currently visible satellites. In Fig.~\ref{Picture2}a, the yellow curved arrows illustrate this dissemination process. In the subsequent rounds ($\beta=1,2,...$) $w^{0}$ is substituted by $w^{\beta}$ while the procedure remains the same.

\subsubsection{Inter-Satellite Dissemination of Local and Partial-Global Models}
In the worker tier, each visible satellite $k$ in the LEO constellation will perform two tasks upon receiving the global model $w^{\beta}$ in the $\beta$-th round. The first is to retrain $w^{\beta}$ on its own data to generate an updated local model $w_{k}^{\beta}$. The second task is to send both $w^{\beta}$ and $w_{k}^{\beta}$ to its next-hop satellite $k'$ via ISL (the dissemination direction is pre-designated, either clockwise or counter-clockwise). If the neighbor $k'$ is invisible to the PS, it will do the same as $k$ but, additionally, also perform a {\em partial aggregation} of its generated local model $w_{k'}^{\beta}$ and the received $w_{k}^{\beta}$ as follows:
\begin{align}\label{eq:partialagg}
    w_{k}^{\beta}= (1-\gamma_{k'}) w_{k}^{\beta} +  \gamma_{k'} w_{k'}^{\beta}
\end{align}
where $\gamma_{k'} = {m_{k'}}/{m}$ is a scaling factor, $m_{k'}$ is the data size of the invisible satellite $k'$ and $m$ is the total data size of the satellites on the same orbit. 
Thus, $w_{k}^{\beta}$ is a partially aggregated model which we call a {\em partial-global model} and will be sent to the next invisible satellite together with the global model $w^{\beta}$. The above process continues until reaching the next visible satellite, which will then transmit the updated $w_{k}^\beta$ to its visible HAP, and stop sending further (while the dissemination {\em originating from itself} to the same direction has started earlier). Fig.~\ref{Picture2}b gives an illustration. If the neighbor $k'$ is visible, this is simply a special case of the above. 


In summary, our inter-satellite model dissemination, together with partial aggregation, ``activates'' all the satellites even though some are invisible, and is thus able to speed up the training process.

\subsubsection{Inter-HAP Dissemination of Partial-Global Models}
Once all the HAPs receive the partial-global models from their respective visible satellites, they will disseminate these partial models in the {\em reverse} direction by starting from the {\em sink} HAP until reaching the {\em source} HAP. Then, the source HAP will aggregate all the received partial models into an updated global model $w^{\beta+1}$, following Section~\ref{sec:aggreg} (Eq. \ref{eq:agg}), and propagate $w^{\beta+1}$ to all the HAPs, following \sref{sec:interhap}.

Algorithm~\ref{algorithm1} summarizes the entire process, and Lines 2-13 correspond to the above three stages.

\subsection{Partial and Full Model Aggregation}\label{sec:aggreg} 
Once the source HAP has collected all the partial-global models from its collaborative HAPs on the server tier, it organizes these models as follows:
\begin{equation} \label{Eqn17}
    \mathcal{U}= \{ \underbrace{\{\mathcal{U}_{h1},.,\mathcal{U}_{{H}}\}_{l_1}}_{S_{l_1}}, \underbrace{\{\mathcal{U}_{h1},.,\mathcal{U}_{{H}}\}_{l_2}}_{S_{l_2}},..,\underbrace{\{\mathcal{U}_{h1},.,\mathcal{U}_{{H}}\}_{L}}_{S_{\textit{L}}}\}
\end{equation}
where $\mathcal{U}$ is a set cover that contains all partial models received from all orbits within the LEO constellation, and $\{\mathcal{U}_{h}\}_{l}$ and $S_{l}$ are subsets of all partial models  collected by HAP {\em h} and all HAPs $H$, respectively, for an orbit {\em l}. If a satellite may be visible to more than one HAP, FedHAP will filter out redundant partial models for each $S_l$ using satellite IDs received by HAPs as metadata. Then, FedHAP results in $\mathcal{U}' = \{S'_{l_1},S'_{l_2},\dots,S'_{L} \}$ where $S'_l$ contains distinct partial models for each orbit $l$. Next, for all orbits $L$ it will check if there is any satellite ID  being left out of $\mathcal U'$\footnote{This scenario could happen when there are no visible satellites or when the generation process of a partial model takes longer than the visibility period of the currently visible satellite of an orbit $l$ during the current global epoch.}.
In such cases, FedHAP will not generate an updated version of the global model. Instead, it will wait until $\mathcal U'$ receives the models of those satellites. 
When $\mathcal U'$ receives all the models from all orbits, 
FedHAP aggregates them as follows:
\begin{equation} \label{eq:agg}
   w^{\beta+1}= \sum_{s'_{l}\in\mathcal {S}'_{L}} \sum_{{{\mathcal {U'}}_h=1}}^{\mathcal {U'}_H} \frac{m_{\mathcal {U'}_h}}{m_{l}} w_{k}^{\beta}
\end{equation}
where $m_{\mathcal {U'}_h}$ is the data size of the partial global model collected by HAP $h$, and $m_{l}$ is the entire data size collected by satellites within an orbit $l$.
Lines 14-18 of Algorithm~\ref{algorithm1} summarize the above processes. Then, the entire process will start over from \sref{sec:interhap}, until convergence. 

\vspace{-.15cm}
\begin{algorithm}
\caption{\small FedHAP Model Dissemination \& Aggregation}\label{algorithm1}
\kwInit{Global epoch $\beta$=0, global model $w^{\beta}$, $\mathcal U_h|_{h=1}^H=\phi$}

\While{Stopping criteria is not met }{
\ForEach(\Comment*[h]{\scriptsize Inter-HAP dissemination of the global model}){$h$ from source to sink HAP}{
  Transmit $w^{\beta}$ to all visible satellites of {\em h}\\
\ForEach(\Comment*[h]{\scriptsize Inter-Sat dissemination of local and partial-global models}){$k\in \mathcal K$ that is visible to $h$ }{  
Retrain $w^{\beta}$ on $k$'s own data to obtain $w_{k}^{\beta}$\\  
\ForEach(\Comment*[h]{\scriptsize Aggregate partial models}){invisible $k'$ between $k$ and $k+1$}{
Retrain $w^{\beta}$ on $k'$'s local data to obtain $w_{k'}^{\beta}$\\
Aggregate $w_{k}^{\beta}$ and $w_{k'}^{\beta}$ using (\ref{eq:partialagg})\\ 
Propagate both $w^{\beta}$ and $w_{k}^{\beta}$ to next $k'$ 
}
$k+1$ transmits $w_{k}^{\beta}$ to its visible HAP\\ 
Update $\mathcal U_{h} \leftarrow  \mathcal U_{h} \cup \{w_{k}^{\beta}\}$ and record all the disseminating Sat IDs
}
}

\ForEach(\Comment*[h]{\scriptsize Inter-HAP Dissemination of partial models}) {$h$ from sink to source HAP}{ 
{Transmit} $\mathcal U_{h}$ to the next neighboring HAP
}
\eIf{Source HAP receives all partial models}{
{Filter} out redundant models from $\mathcal{U}$ \eqref{Eqn17} based on sat ID \\
Aggregate $w^{\beta+1}$ using (\ref{eq:agg})\\
}{{Reschedule} model aggregation to the next epoch}
{$ \beta \leftarrow  \beta+1$}}
  
\end{algorithm}




\vspace{-2mm}
\section{Performance Evaluation}\label{section 3}

\subsection{Experiment setup}
\begin{figure*}[ht]
     \centering
     \begin{subfigure}[b]{0.25\textwidth}
         \centering
         \includegraphics[width=\textwidth]{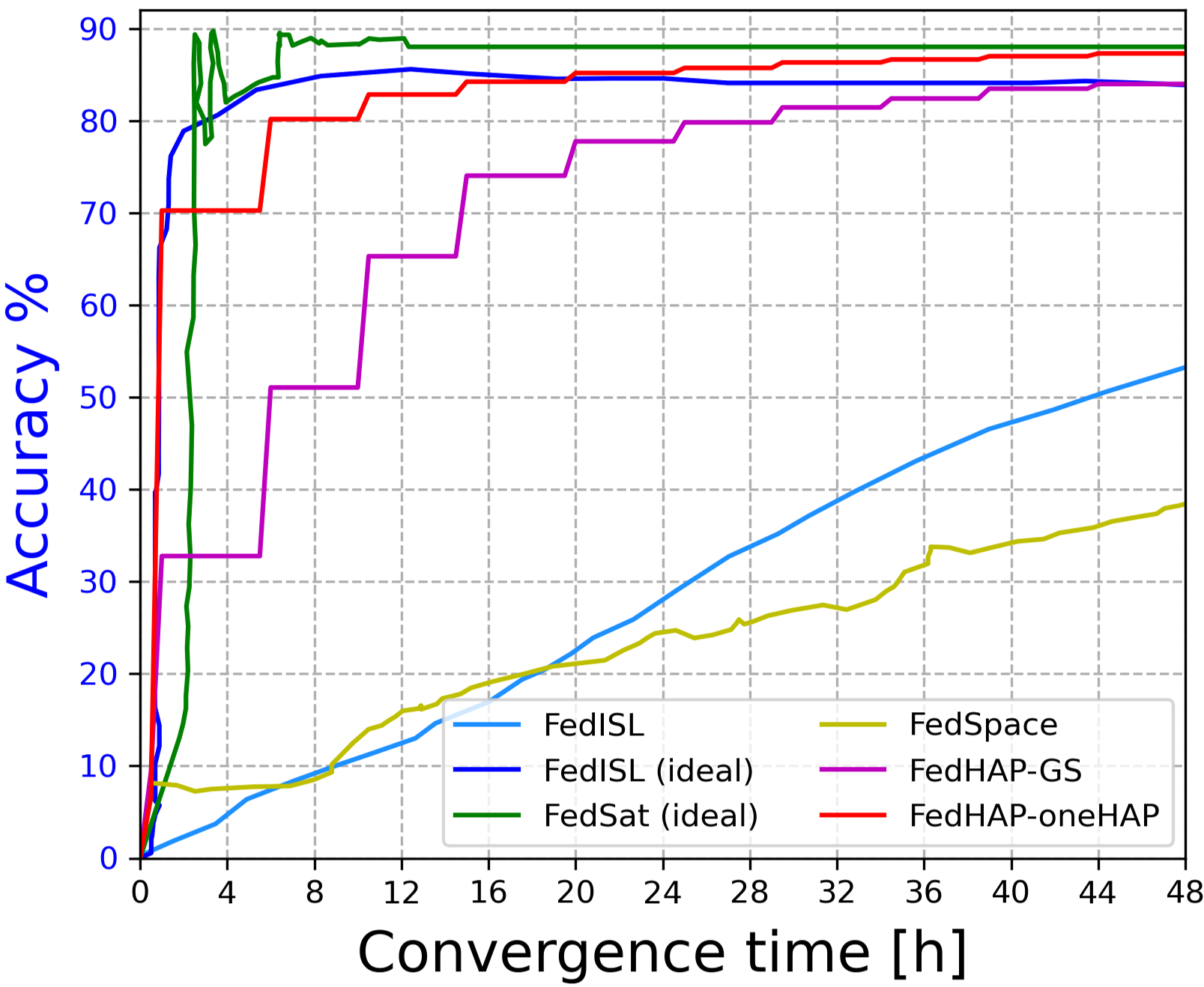}
         \caption{Comparison with baselines.}
     \end{subfigure}  
     \begin{subfigure}[b]{0.24\textwidth}
         \centering
         \includegraphics[width=\textwidth]{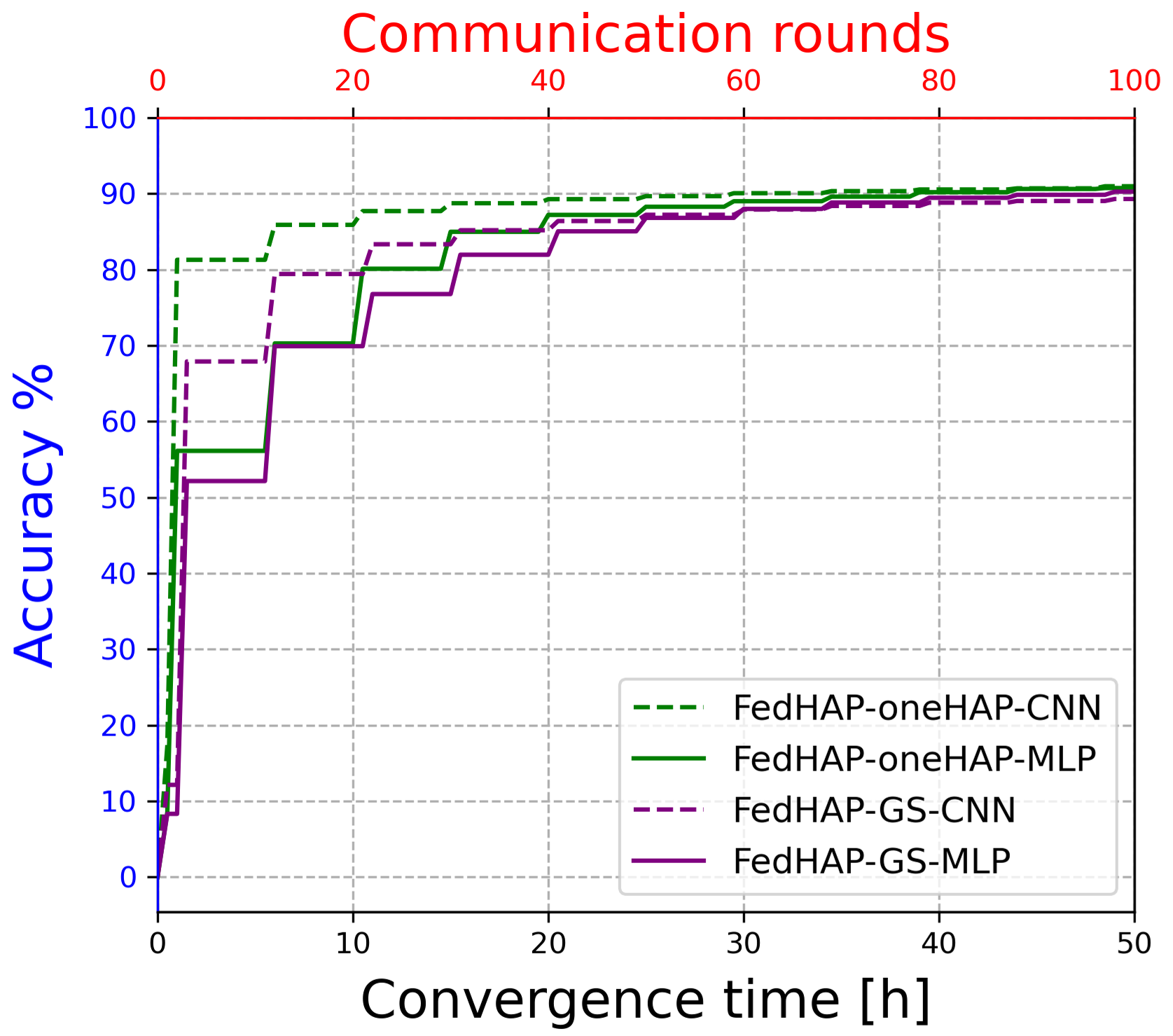}
         \caption{\centering IID data.}
     \end{subfigure}
     \begin{subfigure}[b]{0.24\textwidth}
         \centering
         \includegraphics[width=\textwidth]{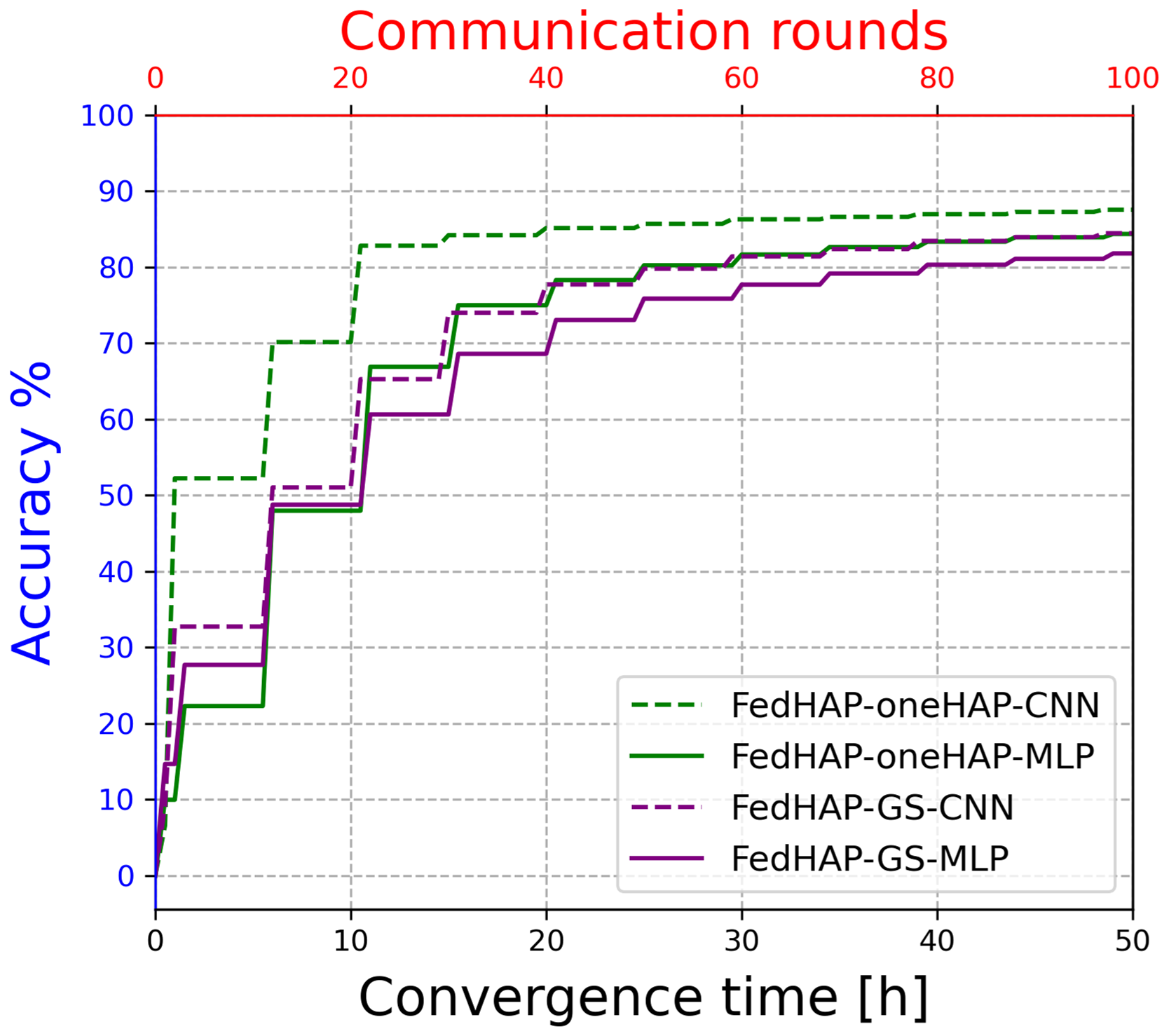}
         \caption{Non-IID data.}
     \end{subfigure}
     \begin{subfigure}[b]{0.24\textwidth}
         \centering
         \includegraphics[width=\textwidth]{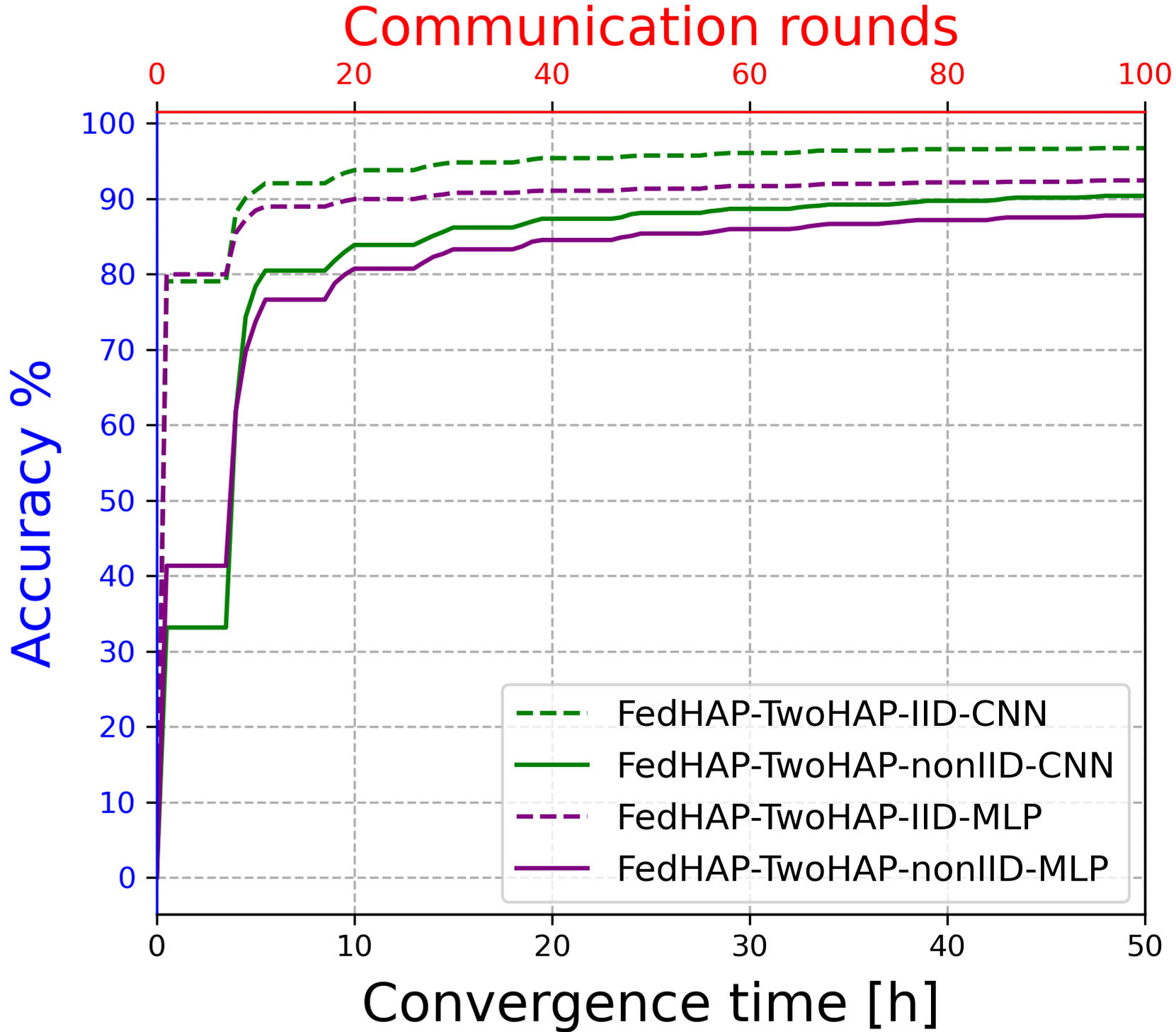}
         \caption{Two HAPs (Rolla and Dallas)}
     \end{subfigure}
        \caption{Evaluation in multiple settings: IID vs. non-IID, CNN vs. MLP, GS vs. HAP, one vs. multiple HAPs.}
        \label{Picture5}
        \vspace{-.4cm}
\end{figure*}
\textbf{LEO Constellation and Communication Links.} We consider a Walker-delta constellation \cite{walker1984satellite} consisting of five orbits, each containing eight LEO satellites equally spaced. Each orbit is at 2000 km above the Earth's surface with an inclination angle of 80 degrees. We consider both single-HAP and multi-HAP scenarios. In the former, the HAP floats above Rolla, Missouri, USA (the same location is also used by the GS in baselines). In the latter, one HAP floats above Rolla and the other above Dallas, Texas, USA. All the HAPs float at 20 km above the Earth's surface with a minimum elevation angle of 10 degrees (the same degree as a typical GS). 
The parameters of communication links discussed in Section \ref{Com_link}, which determines the connectivity between satellites and PSs, are listed in Table \ref{Parameter}, for both FedHAP and the baselines we compare to. Note that all the FSO parameters are chosen in such a way that FSO links behave similarly to RF links, to ensure a fair comparison with baselines (so in reality FedHAP would perform even better). We run each simulation for three days to obtain each set of results.
\begin{table}[h]
 \vspace{-.2cm} 
\setlength{\tabcolsep}{1.1em}
\centering\arraybackslash
\renewcommand{\arraystretch}{1.1}
\caption{Communication Parameters}
\label{Parameter}
\scriptsize
 \begin{tabular}{|p{2.2cm}|p{1.1cm} |p{1.42cm} |} 
 \hline
  \centering Parameter& \centering RF Link& FSO Link \\
 \hline  
  $G$ (sender \& receiver)& 6.98 dBi& same\\
 \hline
 $P_{t}$ (satellite \& PS)& 40 dBm & 10 dBm\\
 \hline
 \centering $f$ & 2.4 GHz& same \\
  \hline 
\centering$T$& 354.81 K& same \\
  \hline 
\centering $R$ & 16 Mb/sec & same  \\ 
 \hline 
 \centering $\mathcal{V}$& \centering --- & 0.021 km/sec \\ 
  \hline 
\end{tabular}
\vspace{-4mm} 
\end{table}

\textbf{Dataset and ML models.} We use the MNIST dataset as is used by most of the FL-Satcom studies in the literature. This dataset contains 70,000 images of handwritten digits of ten classes (0-9), in grayscale with a resolution of 28$\times$28 pixels. We use two models at each satellite for model training: convolutional neural network (CNN) and multi-layer perceptron (MLP). In addition, we investigate both IID and non-IID data distributions. In the IID setting, we shuffle the training samples and randomly distribute them equally among all the satellites, with each having all 10 classes of images. In the non-IID setting, satellites in three orbits have 6 classes (digits 0-5), while satellites in the other two orbits have 4 classes (digits 6-9). Finally, for hyper-parameters, we set the mini-batch size to 32 and the learning rate to $\zeta$ 0.01.

\textbf{Baselines.} We compare FedHAP to the most recent (dated 2022) peer research as reviewed in Section \ref{Intro}, including FedISL \cite{razmi} (synchronous approach), FedSat \cite{razmi2022ground} (asynchronous), and FedSpace \cite{so2022fedspace} (asynchronous).

\subsection{Results}

\textbf{Comparison with State of the Art.}
For a fair comparison, we use only a single HAP or GS and make two versions of FedHAP, FedHAP-GS, and FedHAP-oneHAP, to compare with baselines (later we have a two-HAP version for more extensive evaluation). In FedHAP-GS, everything is the same as what is described in this paper, except that the HAP is replaced by a GS (so it will benefit from our model dissemination  and aggregation).

\begin{table}[h]
\setlength{\tabcolsep}{0.4em}
\centering\arraybackslash
\renewcommand{\arraystretch}{1.1}
\caption{Comparing FedHAP to State of The Art (All non-IID)}\label{compar}
\centering
 \begin{tabular}{|p{2.05cm}|p{1.5cm} |p{1.25cm} |p{3.1cm}| } 

 \hline
  \centering FL model& \centering \scriptsize Accuracy (\%)& \centering \scriptsize Convergence time (h)& Remark \\
 \hline
FedISL \cite{razmi}& 63.74 & 72& GS at arbitrary location\\
 \hline  
\scriptsize FedISL ({\bf ideal}) \cite{razmi} & 82.87 & 3.5& GS at NP or MEO above the equator \\

 \hline
\scriptsize FedSat ({\bf ideal}) \cite{razmi2022ground}& 88.83 & 12 & GS at NP so all satellites visit GS periodically\\
  \hline 
FedSpace \cite{so2022fedspace}& 46.10 & 72 & GS needs satellite raw data\\
  \hline 
FedHAP-GS & 83.94 & 40 &  GS at arbitrary location\ \\ 
 \hline 
     \rowcolor{lime}
FedHAP-oneHAP & \textbf{87.286} & \textbf{30} & HAP at arbitrary location \\
 \hline 
    \rowcolor{lime}
\scriptsize FedHAP-twoHAP & \centering \textbf{80.45}({\bf 89.83}) & \textbf{5 (30)} & HAPs at arbitrary location \\ 
\hline
\end{tabular}
\end{table}
The comparative results are given in Table~\ref{compar} and Fig.~\ref{Picture5}a. We can see that FedHAP-oneHAP converges to an accuracy of 87.3\% in 30 hours (or 80 global epochs) without restriction on HAP locations. This is unlike FedISL \cite{razmi} which requires an ideal setup where GS must be located at the NP, 
yet its accuracy (82.9\%) is still lower than FedHAP. After removing this ideal condition, as shown in the first row, FedISL takes 72 hours (200 global epochs) to converge and the accuracy is only 63.7\%. FedSat \cite{razmi2022ground} assumes the same ideal setup in order to have regular visiting intervals. 
FedSpace~\cite{so2022fedspace} does not assume the ideal setup but converges much slower (72 hours) with low accuracy (46.1\%). Between FedHAP-GS and FedHAP-oneHAP, the latter outperforms the former in terms of both accuracy and convergence time, showing the advantages of using HAPs (recall Section~\ref{Intro}). Nonetheless, even FedHAP-GS performs quite well too, outperforming FedSpace and FedISL without the ideal setup by large margins.

In this set of results, CNN is used as the training model and data are non-IID. More scenarios are evaluated next. 

\textbf{Evaluating FedHAP in more extensive scenarios.}
We investigate FedHAP more thoroughly with more settings, including CNN vs. MLP, IID vs. non-IID data, and single PS vs. multiple PSs. See Fig.~\ref{Picture5}b for the results on IID data. When CNN is used, FedHAP achieves an accuracy of 90.13\% within 20 hours using a single HAP, while FedHAP-GS achieves an accuracy of 89.3\% after 35 hours. When MLP is used instead of CNN, the convergence time does not exhibit a notable change, and the accuracy of both oneHAP and GS versions drops by 1-3\% only. Fig.~\ref{Picture5}c presents the results on non-IID data. It shows that FedHAP is robust to non-IID by performing rather closely to its IID counterpart: it converges in 60 global epochs (30 hours) and achieves an accuracy of 87.3\% with a single HAP, and 80 global epochs (40 hours) and accuracy of 84\% with a GS. This difference in accuracy between HAP and GS is due to the fact that HAP is able to observe more LEO satellites than GS (by about 1-5 based on what we observe in simulations). Switching from CNN to MLP results in marginally reduced accuracy and a 10 hours increase in convergence time. However, this performance is still considerably better than other baseline methods.

In Fig.~\ref{Picture5}d, we present the results for two HAPs, under both IID and non-IID data settings. In the IID case, FedHAP reaches a high accuracy of 92.135\% in only 5 hours and converges to an even higher 96.6\% after 20 hours. In the non-IID case, it achieves an accuracy of 80.452\% within 5 hours and 89.833\% after 30 hours. When CNN is switched to MLP, the performance difference is negligible in the case of non-IID, while the accuracy drops about 5\% (but still above 90\%) in the case of IID, with approximately the same convergence time. Therefore, we can conclude that our proposed inter-satellite/HAP collaboration including model dissemination  and aggregation is effective in accelerating FL convergence and improving model accuracy for LEO constellations.
\vspace{-.2cm}
\section{Conclusion} \label{section 4} 
This paper introduces HAPs into FL-Satcom to orchestrate the iterative learning process and proposes a novel synchronous FL framework, FedHAP, that leverages inter-satellite/HAP collaborations to accelerate FL convergence and improve model accuracy. FedHAP tackles the challenge of highly sporadic and irregular satellite-GS connectivity in LEO constellations using a hierarchical communication architecture, a model dissemination  scheme, and a model aggregation algorithm. Our simulation results demonstrate promising results of FedHAP as compared to the state of the art (5 times faster with an accuracy as high as 97\%), as well as its robustness to non-IID data as is typical in FL-Satcom settings.
\vspace{-.2cm}

\small
\bibliographystyle{IEEEtran}
\bibliography{references.bib}

\end{document}